\documentclass[a4paper,times,10pt,twocolumn]{IEEEtran}

\usepackage[top=4.9cm,bottom=3.7cm,left=1.5cm,right=1.5cm]{geometry}
\usepackage{ICMLC}
\usepackage{times}
\usepackage{graphicx}
\usepackage{indentfirst}
\usepackage{latexsym}
\usepackage{tabularx}
\usepackage{tabularray}
\usepackage{subfigure, subcaption}
\usepackage{balance, xcolor}

\usepackage[margin=8pt,font=footnotesize,labelfont=bf,labelsep=period
]{caption}

\pdfpagewidth=\paperwidth
\pdfpageheight=\paperheight
\begin{document}

\title{FEDERATED LEARNING FOR LARGE-SCALE CLOUD ROBOTIC MANIPULATION: OPPORTUNITIES AND CHALLENGES}  

\author{\bf{\normalsize{OBAIDULLAH ZALAND${^1}$, CHANH NGUYEN${^1}$, FLORIAN T. POKORNY${^2}$, MONOWAR BHUYAN${^1}$}}\\ 
\\
\normalsize{$^1$Department of Computing Science, Ume\r{a} University, Ume\r{a} SE-90187, Sweden}\\
\normalsize{$^2$Division of Robotics, Perception and Learning, KTH Royal Institute of Technology, Sweden} \\
\normalsize{E-MAIL: \{ozaland, chanh, monowar\}@cs.umu.se, fpokorny@kth.se }\\
\\}

\maketitle 

\begin{abstract}
Federated Learning (FL) is an emerging distributed machine learning paradigm, where the collaborative training of a model involves dynamic participation of devices to achieve broad objectives. In contrast, classical machine learning (ML) typically requires data to be located on-premises for training, whereas FL leverages numerous user devices to train a shared global model without the need to share private data. Current robotic manipulation tasks are constrained by the individual capabilities and speed of robots due to limited low-latency computing resources. Consequently, the concept of cloud robotics has emerged, allowing robotic applications to harness the flexibility and reliability of computing resources, effectively alleviating their computational demands across the cloud-edge continuum. Undoubtedly, within this distributed computing context, as exemplified in cloud robotic manipulation scenarios, FL offers manifold advantages while also presenting several challenges and opportunities. In this paper, we present fundamental concepts of FL and their connection to cloud robotic manipulation. Additionally, we envision the opportunities and challenges associated with realizing efficient and reliable cloud robotic manipulation at scale through FL, where researchers adopt to design and verify FL models in either centralized or decentralized settings.
 
\end{abstract}

    
\section{Introduction}


\begin{figure*}\label{fig:graspingmodels}
    \centering
    \subfigure[]{\includegraphics[width=0.4\textwidth]{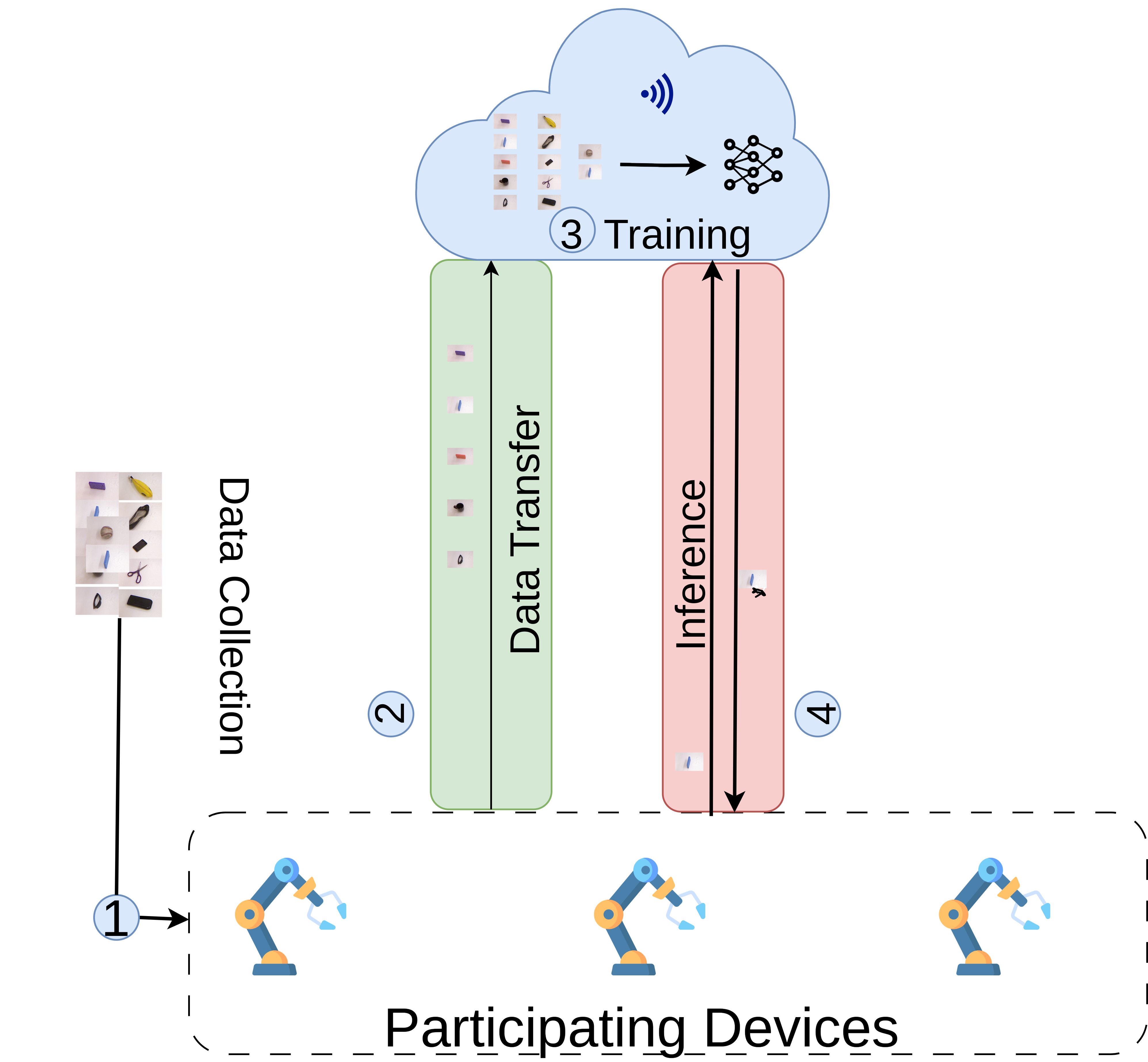}}
    \subfigure[]{\includegraphics[width=0.4\textwidth]{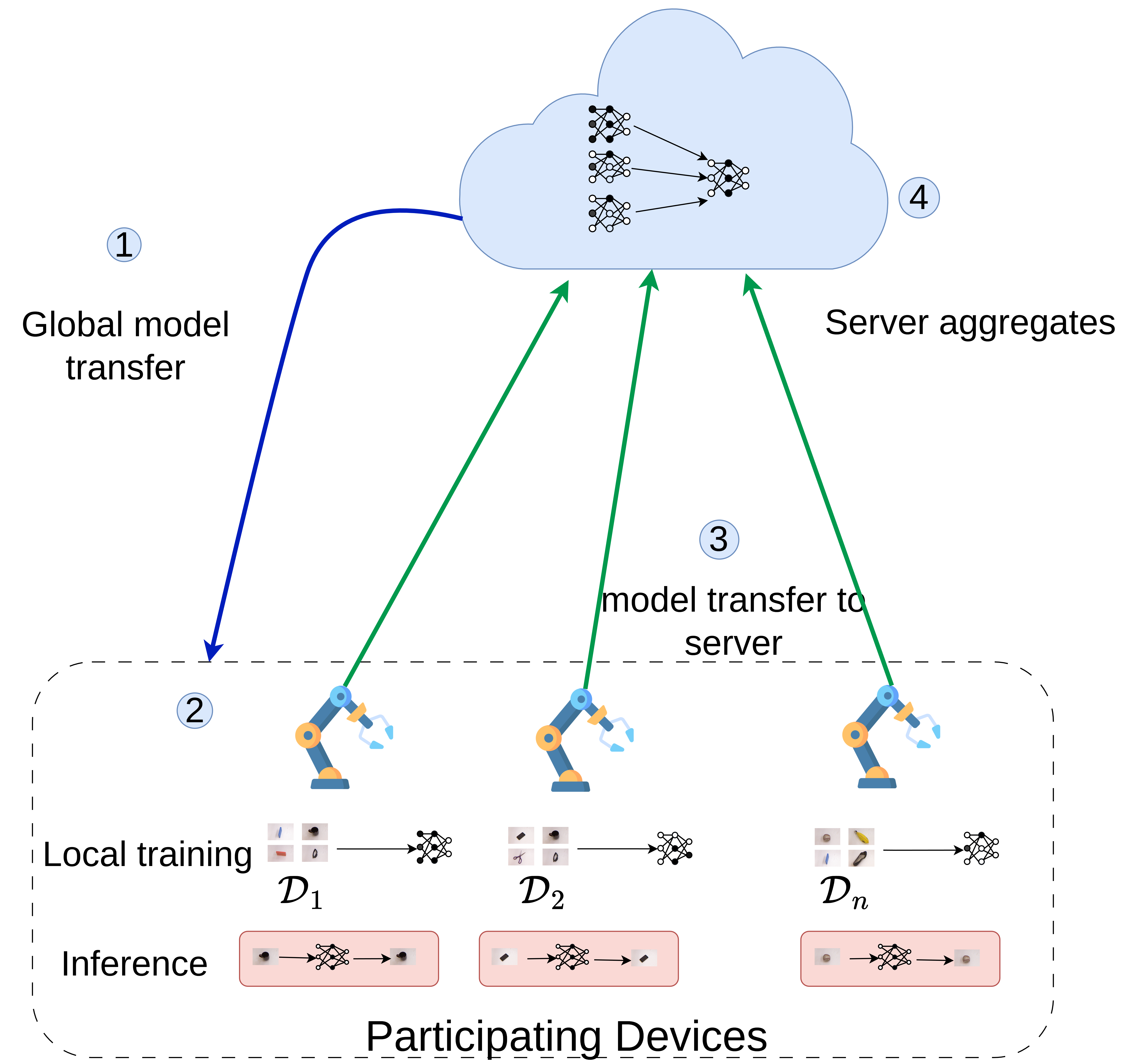}}
    \caption{Centralized vs. federated cloud robotic setups.} 
    \label{fig:FL}
\end{figure*}

The increasing deployment of robots in various domains has resulted in large volumes of data. Consequently, there has been a rising need for powerful machine-learning models that can handle and utilize them. While the development of distributed machine learning paradigms has presented a promising venue, the challenges of \textit{privacy} and \textit{data ownership} remained unsolved amidst growing data privacy regulations \cite{european_commission_regulation_2016}. Federated Learning (FL) is an emerging machine learning paradigm that harnesses the computational power of user devices and utilizes distributed data for model training while prioritizing data privacy. Robotic manipulation is a compelling application for FL \cite{liu2021federated}, where robots are trained on private data to get a high-performing personalized model that can help, for example, elderly home-bound people in different day-to-day household tasks.
As the demand for sophisticated models trained on vast and diverse datasets has grown amidst privacy concerns, federated learning (FL) \cite{zaland2025one} has evolved as a training methodology for machine learning applications. In its simplest form, an FL setup consists of a server and $n$ participating devices, where the server initializes a global model and shares it with participating devices. Each participating device trains the model with its local data and sends it back to the server, where the server makes an aggregation in each round and repeats this for the next round until the model converges.  
 
Robotics has risen as a transformative multidisciplinary field that revolves around studying intelligent machines' design, creation, and operation.  These intelligent machines (robots) are capable of performing tasks autonomously and can usually \textit{sense}, \textit{plan}, and \textit{act} using, for example, manipulation. In order to plan in complex environments, robots require adequate computational capabilities, and to keep robots agile, computation or storage can be offloaded to the cloud (cloud robotics) \cite{kehoe2015survey}. Cloud integration offers a scalable, collaborative, and maintainable infrastructure for deploying robotic fleets (groups of robots aiming to do similar tasks). While cloud offloading seems binary, the robots can decide to plan smaller tasks locally to reduce latency and offload when greater computational capability is required.  The rising deployment of robotic systems can be attributed to their ability to manipulate complex and unpredictable environments. Manipulation is often defined as ``making an intentional change to the environment''\cite{mason2018toward}. The intentional change can be made using pushing, grasping, picking, rearranging, etc. Robots need to manipulate their surroundings in dynamic environments. In order to excel at these manipulation tasks, robots need large volumes of manipulation task-specific data, which is costly and time-consuming for a single robot to collect and enables massive-scale training. 

Federated learning is key and has the potential to address the challenges in cloud robotics and robotic manipulation. Several inherent challenges exist in FL itself, such as statistical heterogeneity, system heterogeneity, communication, privacy, and security. However, a few of them are important to explore when it comes to training a large number of robots. 
Additionally, adaptability becomes increasingly important when interacting with the environment, and FL-based collaboration can enhance robots' capability. Decentralized learning ensures that robots adapt to diverse scenarios while keeping their data private. Additionally, as robots are generally deployed in fleets \cite{sikand2021robofleet}, FL can enable collaborative decision-making by utilizing each robot's individual insights and experiences to improve the fleet's overall perception and manipulation capabilities without raw data transfer.


Despite the fact that the application of ML has been extensively studied in robotic manipulation, FL for robotic systems has been studied at a very surface level. Applying federated learning to robotics and cloud robotics brings up unique and unforeseen challenges. As per our knowledge, no existing literature studies the aforementioned fusion and discusses the prevalent challenges and possible solutions. We believe that the intersection of FL and cloud robotic manipulation would pave the way for trustful, private, robust, scalable, and efficient robotic systems and could speed up the deployment of robotic fleets in various application domains. This work outlines the existing literature in the intersection of cloud robotics and federated learning, alongside  outlining existing challenges. Further, this work provides opportunities and future research directions to improve the existing landscape of federated learning enabled cloud robotic applications. 


\section{Federated Learning}

As the need for large training datasets in machine learning (ML) applications grew, distributed ML paradigms were increasingly applied to distribute the training over multiple systems while learning a single machine learning model \cite{xing2015petuum}. Distributed ML algorithms require \textit{full ownership} of the data \cite{shen2022distributed}, and an improved privacy-preserving collaborative learning paradigm was required in the age of increased data privacy regulations. Federated Learning is a machine learning training paradigm where the participating devices collaborate to train a joint model without explicitly sharing their data. The local storage of data helps FL tackle privacy challenges while enabling collaborative learning. FL systems can be divided into centralized FL and decentralized FL systems. Similar to centralized ML, centralized FL requires a server to coordinate the learning process, while decentralized FL is based on peer-to-peer communication among the participating devices.

The centralized FL process can be divided into three broad sections: i) model initialization, ii) local training, and iii) server aggregation. In the model initialization phase, the server initializes a global model that is communicated to all or a subset of participating devices. The device selection in each communication round might help avoid straggler devices and improve overall fairness. As a subset of devices is selected, they receive the global model from the server, train the global model on their local data for a fixed number of epochs, and send the model back to the server. The server aggregates the received models and consequently starts the next iteration. In the simplest form, federated learning uses FedAvg \cite{konevcny2016federated}. In FedAvg, the server calculates a weighted average of the received models, where the clients' weights are the proportion of the total training data they hold. Local models are aggregated in the server according to equation \ref{eq:FedAvg}, where $\omega_t$ represents global model weights at time $t$ and is calculated as a weighted average of local model weights for all devices at time $t-1$. 

\begin{equation} \label{eq:FedAvg}
    \omega^t = \sum_{c \in \mathcal{C}_m} {\frac{n_c}{N}} \omega_c^{t-1}
\end{equation}
while FedAvg is the simplest aggregation solution in an FL environment, complex aggregation functions such as FedProx \cite{li2020federated} are increasingly used, as they can alleviate the intrinsic system heterogeneity of FL environments better.

\textbf{Federated Learning Approaches:} Based on data distribution, FL can be divided into horizontal and vertical federated learning. Horizontal federated learning (HFL) is suitable for FL scenarios where the features among the participating devices' datasets are the same and the data instances differ. In vertical federated learning (VFL), the participating devices hold different features of the same instances. The devices share the same sample space but provide different information about each subject. In VFL, unlike HFL, the devices may not share the model parameters and only share \textit{intermediate results} and keep their local model private \cite{wang2025pravfed}. Hybrid approaches can utilize both setups concurrently.

\section{Cloud Robotics -- The Intersection of Cloud-Edge Continuum and Robotics}

\begin{figure}[!ht]
    \centering
    \includegraphics[width=0.42\textwidth]{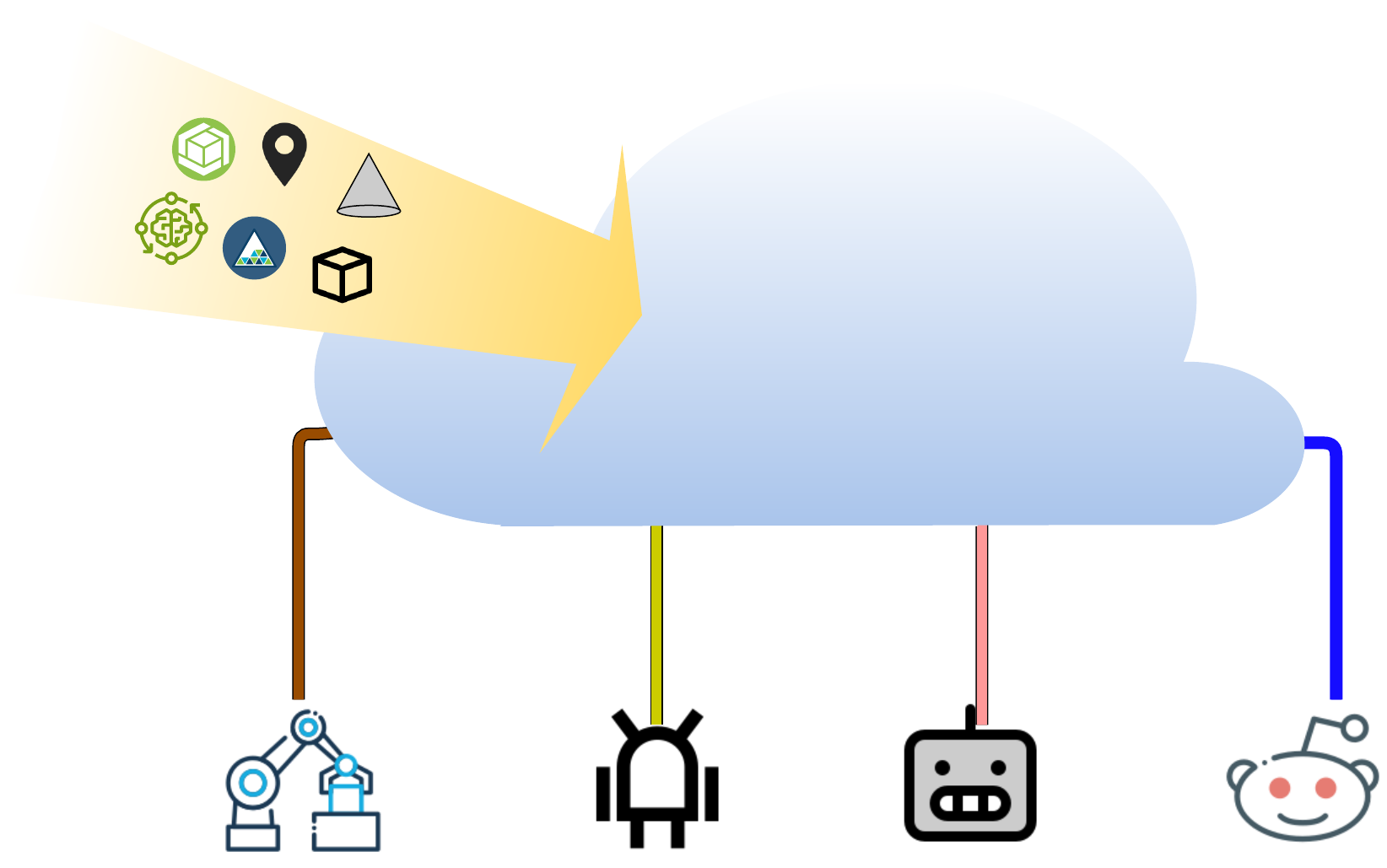}
    \caption{Cloud robotics ecosystem: robots connected to the Cloud for perception, understanding, reaction, and knowledge sharing.}
    \label{fig:CR}
\end{figure}

In this section, we explore the convergence of cloud computing and edge computing in what is known as the \textit{Cloud-Edge Continuum}, highlighting their combined potential to establish an extensive network of cloud-connected robots.
\subsection{The Cloud-Edge Continuum}
\textit{Cloud computing}, since its inception over two decades ago, has fundamentally reshaped the landscape of information technology. It has transcended from a novel concept to a ubiquitous infrastructure for deploying applications worldwide. A key driver behind cloud computing's pervasive adoption lies in its remarkable \textit{resource elasticity}, which enables organizations to adjust computing resources in response to demand dynamically, optimizing efficiency and scalability.
More recently, the advancement of sensor technologies and IoT applications has brought forth fresh challenges characterized by stringent demands for low latency and high bandwidth. This has given rise to a novel distributed computing infrastructure, where computing and storage capabilities are strategically placed close to the data sources or end-users at the network's edge. This transformative paradigm is referred to as \textit{Edge computing}~\cite{shi2016edge}.
Edge computing provides resources within edge data centers, often known as ``cloudlets'', resembling a compact cloud infrastructure located in close proximity. Experimental studies have validated and quantified the manifold benefits of edge computing~\cite{satyanarayanan2019seminal}. These advantages encompass substantial improvements in response times and extended battery life, achieved through the offloading of computation from mobile devices to nearby servers instead of the distant cloud. Furthermore, there is a substantial reduction in the demand for ingress bandwidth due to the processing of high-data-rate sensor streams at edge locations close to the data source. Additionally, edge servers positioned in proximity to end-users can function as privacy barriers, granting users dynamic and selective control over the release of sensitive sensor data to the cloud.

The transition between centralized and decentralized computing has oscillated over the course of computing history, mirroring the dynamic demands arising from emerging applications and their core functional requirements. Leveraging resources from the combined paradigms of edge computing and cloud computing, known as the \textit{Cloud-Edge continuum}~\cite{rosendo2022distributed}, not only enhances availability and reliability but also facilitates the flexible deployment of application services while accommodating their multi-objective requirements.
\subsection{Cloud Robotics}
A robotics system consists of numerous programmable and controller components, each operating both independently and cooperatively towards a shared ultimate objective.
To illustrate, consider a robot gripper designed for use in a factory setting, tasked with picking up and placing items. 
This system incorporates a range of cognitive components, including those powered by computer vision models that facilitate object recognition and environmental assessment. 
Furthermore, it features motion planning components that analyze and determine feasible or optimal routes for the robot to navigate from its initial position to the desired goal state. Another important component is the generation of maps to create a comprehensive layout of the workstation. 
Lastly, a kinematics controller is in place to oversee the physical aspects of the robot, such as its orientation, position, and the coordination of its arm and fingers, ensuring precise and efficient control.
Each of these components necessitates diverse resource allocation capabilities, including computing power, storage, and network resources, to attain optimal performance. 
In this scenario, the previously mentioned diversity in resource allocation capabilities provided by the cloud-edge continuum makes it an appealing platform for deploying the robotics system.

Indeed, the concept of utilizing substantial computational resources from cloud data centers for robotics applications was initially introduced over a decade ago, referred to as \textit{cloud robotics}~\cite{kuffner2010cloud}. 
This approach not only provides robots with access to significant computational power and extensive data resources but also facilitates the exchange of collective knowledge and expertise among robots. Consequently, robots can offload computationally intensive tasks and rapidly acquire new skills from the cloud. Figure~\ref{fig:CR} illustrates a visual representation of the cloud robotics concept.
More recently, the cost-effective advantages of 3D printing, prototyping, and open-source robot component designs have opened up opportunities for conducting large-scale robotics experiments. Consequently, cloud robotics has gained significant attention in both academic and industrial domains~\cite{robomaker}, particularly within the realm of large-scale robotic manipulation. 
Several research endeavors are dedicated to the deployment of bridge platforms, aimed at unifying the deployment of robotic applications in a hybrid computing environment spanning robots, edge computing, and cloud computing. Notable examples of such initiatives include Rapyuta~\cite{mohanarajah2014rapyuta}, FogROS~\cite{ichnowski2023fogros}, and KubeROS~\cite{zhang2023kuberos}.
Moreover, certain initiatives have embarked on exploring the implications of \textit{large-scale data collection} across a variety of robotic platforms~\cite{noh2023x}. This underscores the critical importance of the generation of extensive datasets for real-world robotic tasks.

In the domain of large-scale cloud robotics, as these technological revolutions were unfolding, concerns related to \textit{data security and end-user privacy} have become inevitable research focal points. 
FL stands out as a promising solution for preserving privacy during edge-based deep learning. 
Its inherent strength, as mentioned earlier, lies in its distributed approach, wherein it learns from isolated data islands and exclusively shares model updates.
To better appreciate the advantages of FL in this context, consider a scenario where teleoperated robots assist with healthcare tasks in a hospital.
Here, ensuring the protection of data security and end-user privacy becomes paramount. The robot is responsible for collecting sensitive patient data and relies on remote cloud-based control. In this scenario, FL \textit{enables local data processing} directly on the robot itself, ensuring that sensitive information remains securely stored on the device. This approach enables machine learning models to \textit{collaboratively train across multiple robots or facilities} without exposing individual patient data, thereby enhancing both data security and privacy. Furthermore, the \textit{implementation of secure communication protocols} guarantees the safe transmission of data between the robot and the cloud-based control interface. In this comprehensive manner, FL effectively addresses the challenges associated with data security and end-user privacy, particularly in sensitive domains where robots are deployed.


\section{Early Efforts in Leveraging Federated Learning for Cloud Robotics}

The application of FL to robotics and cloud robotics, in general, has been studied at a superficial level, primarily due to complexities in these resource-constrained and dynamic environments. The challenges of data and device heterogeneity and varying computational capabilities have limited FL adaptation in such autonomous systems. Despite the obstacles, FL can potentially mitigate several issues in cloud robotics. In cloud robotics, where privacy is of utmost importance, federated learning (FL) enables local model training by keeping data private on individual robots. It can additionally foster interoperability between multiple robots to mitigate heterogeneity. In short, while FL has limited applications in cloud robotics, it can emerge as a solution to various challenges in cloud robotic manipulation. This section focuses on current works that utilize FL in robotic and cloud robotic systems, as well as existing FL approaches for these systems.

Given the intrinsic need for robotic systems to comprehend and manipulate complex and dynamic environments, different FL techniques are tailored to specific applications. Machine learning techniques, including reinforcement learning (RL) and, specifically, deep reinforcement learning (DRL) \cite{li2017deep} strategies, have been adapted in robotic manipulation applications due to their adaptability to unseen environments. Swarm Deep reinforcement learning (SDRL) \cite{zhu2022swarm} employs a decentralized, federated reinforcement learning (FRL) setup, where the robots share their local actor-critic models by interacting with their environment between them to train a shared global model. To improve the overall security of the system, SDRL utilizes blockchain. The results indicate faster model convergence as the number of devices increased. The peer-to-peer communication in SDRL, while preserving privacy and security, may lead to communication overload on the network. 

FLDDPG \cite{na2023federated} has studied FRL for swarm robotics for collective navigation, pursuing a similar approach to SDRL, where they collected local navigation data and trained local actor-critic models before aggregating a global model. Their experiments conclude that FL-based system not only overperforms centralized DDPG approaches; they significantly reduce \textit{communication costs extensively} and task execution times. They further argue that FLDDPG improves robustness and generalization in a new environment following real-world experiments. 

Privacy preserved asynchronous FL (PPAFL) \cite{10105168} applies peer-to-peer FL to mobile robotic swarms in 5G and beyond networks. PPAFL assigns clients to temporary clusters based on a \textit{reputation evaluation algorithm}. Clients communicate their local models within these virtual clusters to learn a joint model collaboratively. As soon as the FL phase is terminated, clusters are resolved. The authors argue that decentralized FL improves resilience and security compared to centralized approaches, as it eliminates a central point of failure, alongside distributing communication load over the network. While the inter-cluster communication in PPAFL can help with task-specific models, the lack of intra-cluster communication may lead to a lack of generalization. Additionally, as cluster members increase, peer-to-peer communication may lead to network overload inside the virtual cluster.

Further works have incorporated federated transfer and imitation learning in various robotic applications. Federated imitation learning \cite{liu2020federated} proposes a framework where different participating devices learn different tasks, and a fusion algorithm in the cloud aggregates the task-specific knowledge to form a novel model that can be used in all of the participating scenarios. This can improve fleet generalization when single robots have access to single-task data in a multitasking environment. LFRL \cite{liu2019lifelong} proposes an architecture for lifelong or continual learning in robotic environments. In the proposed architecture, a single global model $1G$ is initialized and shared with all the robots. As soon as a robot shares its local model, it is aggregated to the global model, and a new global model $2G$ is formed and shared with all the participating devices. The model aggregation occurs whenever a participating robot shares its local model with the server. While this architecture is communication-intensive, its asynchronous nature adapts well to cloud robotics architecture. Furthermore, this work provided a starting point for the fusion of cloud robotics and FL.

Studies incorporating FL into robotic and cloud robotic systems have been limited, but FL holds immense potential in the context of cloud robotic systems driven by its distinct advantages over applying ML algorithms. The decentralized and privacy-preserved nature of FL couples well with the collaborative nature of cloud robotics. Hence, more studies are required to investigate the intersection and design more robust and efficient algorithms.

\section{Current Research Challenges}
While integrating FL and cloud robotics can bring forth many opportunities, it also introduces a set of nuanced challenges that need careful contemplation. In this section, we set forth some of the challenges that can arise when applying FL to cloud robotics and underline the significance of addressing these challenges for secure robotic manipulation. 

\subsection{Communication and Latency}
While cloud computing plays a pivotal role in augmenting the computational capabilities of robotic systems, it concurrently introduces high communication costs. The significantly large size of robotic data and the intensive data transfer between robots and the cloud result in an intensified load on the communication channels, resulting in latency issues. While the introduction of FL can substantially reduce the data transfer between the robots and the cloud, the considerable size of robotic manipulation networks and the steadily increasing size of robotic fleets may sustain a higher burden on the communication channels. The burden on the communication channel manifests as delays in cloud-offloaded tasks and may further deteriorate the FL system by introducing stragglers and subsequent aggregation delays. Hence, there is a need to develop adaptive FL algorithms that can efficiently utilize the network bandwidth while minimizing the communication load to prevent delays for robotic systems.

\subsection{Heterogeneity}
Device and statistical heterogeneity are intrinsic to all FL applications. Unlike distributed machine learning, FL applications deal with non-independent and identically distributed (non-iid) data, e.g., statistical heterogeneity. Statistical heterogeneity encompasses both label and quantity skew among the participating parties \cite{li2022federated}. Additionally, the participating parties may differ with regard to their architecture, computing, and network capabilities, e.g., device heterogeneity. The presence of device heterogeneity is intensified as not all participating parties engage in every training round. Additionally, as we apply FL to robotics, the inherent heterogeneity in robotics data with regard to different environments adds a new layer of complexity. Robots are deployed in different environments, executing distinct tasks; hence, the data generated by their sensors vary vastly. A good example can be cooperative heterogeneous multi-robot systems (CH-MAS) that function as a robotic fleet while carrying out different tasks \cite{rizk2019cooperative}. 


\subsection{Security and Privacy}
Security and privacy are critical for cloud robotic systems. The integration of the cloud into robotics introduces new challenges to robotic security, including data breaches and malicious task execution. In cloud robotic systems, the robots may use the cloud for storage alongside communication \cite{dzung2005security}. Robotic data is sensitive in its nature and requires security at both storage and communication levels. For example, a healthcare robot \cite{riek2017healthcare} may store and transmit patients and their diagnosis information or a home monitor robot may collect inside videos and images of a house. Additionally, security influences trust and safety. A malicious task delegation can affect the safety of the robot and its neighborhood. FL helps reduce data privacy concerns through local training and keeping local data on-device, but critical information can still be revealed to the server or a third-party intruder from the model parameters. While approaches such as differential privacy and integral privacy improve data privacy in FL \cite{sun2020ldp}, they often deteriorate model efficiency performance.  

While FL can provide data privacy by enabling collaboration with local data storage, other aspects of the system, including the cloud and communication channels, need proper security mechanisms to prevent privacy leakage from the communicated model parameters. Hence, the intersection requires secure networking, safety by design, authorization, and verification to retain trust and safety \cite{cerrudo2017hacking}.

\subsection{Limited Resources}
Edge and cloud complement each other in addressing the challenges of limited resources in cloud robotics. In order to keep the robots mobile and easy to navigate, on-robot resources are usually limited. Cloud computing certainly helps to provide the power required for intensive tasks but with added communication delays. We can deploy decision-making algorithms to decide on task execution either locally or on the cloud based on the task requirements. 

\subsection{Energy Efficiency}
Energy efficiency poses a significant challenge in the realm of FL within cloud robotics due to the inherent limitations of robotic hardware. Robots often rely on finite power resources, such as batteries, and must operate in real-time, dynamic environments. FL, which involves computational tasks and wireless communication with cloud servers, can be power-intensive, jeopardizing the robot's operational longevity and real-time responsiveness. Achieving a balance between efficient learning and preserving energy resources is essential for successful deployments, demanding innovative algorithms and optimizations tailored to the unique energy constraints of robotic platforms.
Some initial research has investigated the trade-off between learning delay and energy consumption in wireless communication networks~\cite{zhou2023resource}. Furthermore, there have been proposals to reduce the overall energy consumption of FL processes~\cite{kaleem2024hybrid}. While these studies may focus on different perspectives and applications, they can serve as foundational insights for integrating FL in the cloud robotics domain.

\section{Future Research Directions}
\subsection{Clustered FL}
Robotic fleets carry out different manipulation tasks. A robotic fleet may consist of robots packaging, organizing, and carrying boxes. This multi-task nature of robotics can be utilized with clustered FL \cite{xiao2021clustered}. Clustered FL-organized users execute similar tasks to clusters, where they learn a global model alongside optimized model parameters for the users in the same cluster. Robots carrying out similar tasks may be grouped together to learn a task-specific cluster-optimized personalized model alongside a global model. Clustered FL can help robots specialize in a specific task while being able to perform all tasks allocated to the fleet. Clustered FL can further help distribute the communication load, as all the models would not be sent directly to the server. It can also

\subsection{Integration of LLMs}
The research landscape on large language models (LLMs) has grown immensely over the last few years. Models such as GPT-4, Roberta, and Bard are ubiquitous in every aspect of today's society. LLMs can understand and generate human-like text; hence, robotics integration can help robots understand, reason, and manipulate their surroundings better. They can further enhance human-robot interaction and streamline communication between robots with diverse architectures. Additionally, integrating multi-model LLMs that can understand visual input alongside textual instructions can enable robots to understand the task better. 

There is a need for further research to develop multi-model resource-efficient federated LLMs for robotic manipulation. These multi-model LLMs can also be used as foundational models for robotic manipulation tasks \cite{xu2024survey}. 

\subsection{Responsible FL}
As we increasingly rely on ML technologies, concerns over trustworthiness, interpretability, and fairness in ML have been raised. Responsible ML discusses developing fair, accountable, explainable, and trustworthy ML algorithms. Responsible FL extends the idea of trustworthiness to federated environments. With the increasing number of federated robotic fleets capable of manipulation deployed in complex environments, safety, decision-making transparency, and ethical considerations become unavoidable. This requires the development of FL systems capable of decision traceability, handling biases, and considerable care toward safety, privacy, and security. 

\subsection{FL for interoperability}
Robotic fleets are heterogeneous in nature, with diverse environments around them. Seamless interactions and collaboration require interoperability among robots. Interoperable FL \cite{roschewitz2021ifedavg} can help mitigate data level heterogeneity among robotic clients. It can also enable knowledge sharing among diverse domains, resulting in better-generalized models. The interoperable FL for these systems should be developed while keeping in consideration the operational divergence, scalability, and sensor disparities in cloud robotic systems while keeping privacy and security intact. 

\subsection{Advancing security and trust}
Trust can studied from two different aspects in FL-based cloud robotic systems. The system perspective and the human perspective \cite{khan2010establishing}. While trust from a human standpoint is critical, we believe that FL-based cloud robotic systems need to establish trust between various system components. As we integrate FL and cloud with robotics, it is necessary that the cloud can trust the participating parties and vice-versa. The cloud can provide control for the robotic system, while FL can enhance the sense of ownership and privacy among them. Trust and security have a reciprocal relationship. While establishing a secure system results in the trust of the individual parties, it can also act as a precursor, as the individual parties' trust establishment provides the basis for a secure system. We believe that federated learning can help with trustworthiness in cloud robotics, and we need further research to establish the challenges concerning security in trust in FL-based cloud robotic systems.

\section{Concluding Remarks}
In this paper, we discussed the application of FL to cloud robotic systems. We evaluated the current literature on this topic, outlined current challenges, and provided future research directions. While we believe that FL can enhance privacy in cloud robotic systems, we encourage further research to mitigate heterogeneity, improve communication efficacy, and establish trust in security in FL-based cloud robotic systems. 

\balance
\bibliography{bibliography}
\end{document}